\documentclass{bmvc2k}
\usepackage{enumitem}
\usepackage{amsmath}
\usepackage{float}
\usepackage{hyperref}
\newcommand\fnurl[2]{%
\href{#2}{#1}\footnote{\url{#2}}%
}
\usepackage{setspace}

\title{Persuasive Faces: Generating Faces in Advertisements}

\newcommand{\ct}{\textcolor{black}}

\addauthor{Christopher Thomas}{chris@cs.pitt.edu}{1}
\addauthor{Adriana Kovashka}{kovashka@cs.pitt.edu}{1}

\addinstitution{
Department of Computer Science\\
University of Pittsburgh\\
Pittsburgh, PA USA
}

\runninghead{Thomas and Kovashka}{Persuasive Faces: Generating Faces in Ads}

\begin{document}

\maketitle

\begin{abstract}
In this paper, we examine the visual variability of objects across different ad categories, i.e. what causes an advertisement to be visually persuasive. 
We focus on modeling and generating \emph{faces} which appear to come from different types of ads. For example, if faces in beauty ads tend to be women wearing lipstick, a generative model should portray this distinct visual appearance. Training generative models which capture such category-specific differences is challenging because of the highly diverse appearance of faces in ads and the relatively \ct{limited amount of available training data.} To address these problems, we propose a conditional variational autoencoder which makes use of predicted semantic attributes and facial expressions as a supervisory signal when training. We show how our model can be used to produce visually distinct faces which appear to be from a fixed ad topic category. Our human studies and quantitative and qualitative experiments confirm that our method greatly outperforms a variety of baselines, including two variations of a state-of-the-art generative adversarial network, for transforming faces to be more ad-category appropriate. Finally, we show preliminary generation results for other types of objects, conditioned on an ad topic.
\end{abstract}

\section{Introduction}
\label{sec:intro}

\begin{figure}[t]
\label{fig:concept}
  \centering
\includegraphics[width=1\textwidth]{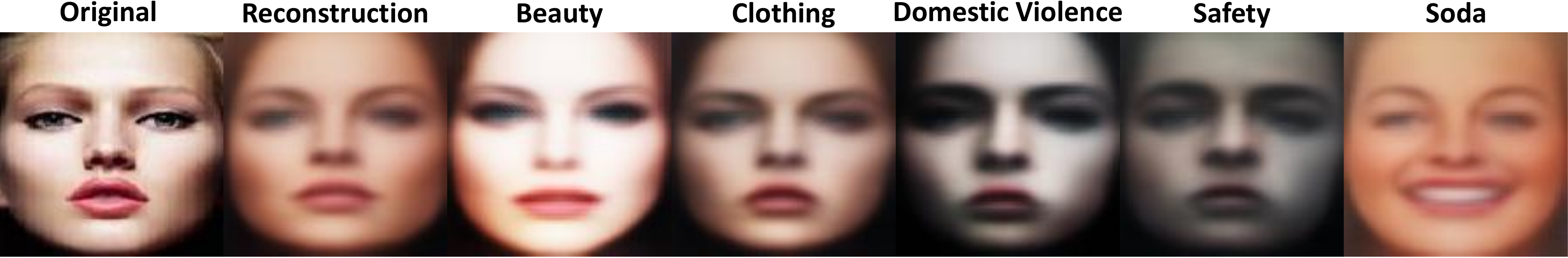}
\vspace{-2em}
\caption{We transform faces so they appear more persuasive and appropriate for particular ad categories. We show an original face on the left, followed by our method's reconstruction without any transformation. We then show the face transformed according to five types of ads. Notice how the beauty face contains heavy make-up, the domestic violence face is sad and possibly bruised, the safety face is somewhat masculine, and the soda face is happy.}
\vspace{-1em}
\end{figure}

Advertisements are persuasive tools that affect people's habits and decisions. They often advertise products and establishments, such as cosmetics and beauty, clothing, alcohol, automobiles, or restaurants. However, they can also be public service announcements that aim to educate the public about important social issues, such as domestic violence or environmental protection. Many topics advertised by ads contain distinctive objects, e.g. the most common object in car ads might be cars, bottles for alcohol ads, and faces for cosmetic ads. There is more to ads than what objects they contain, however. It is \emph{how} objects are portrayed that makes an ad persuasive. For example, faces frequently appear in both beauty and domestic violence ads but their portrayal is vastly different.

What is it that makes a face become a beauty ad or a domestic violence prevention ad?
This is what we set out to discover in this study. 
We first analyze the distribution of objects in common ad topics (beauty, soda, domestic violence, safety, etc.) 
Based on the object distributions, we select to model the appearance of faces, since faces are the most frequent object across all ad categories and have the most distinctive appearance per category. 
We then learn a generative model capable of transforming faces into each ad topic. 
Because ads are rarer than general images, we must work with a sparser dataset than modern generative approaches usually assume. Thus, we propose a method for \emph{transferring  knowledge} from faces in other datasets, in order to mimic the variability of faces in the ads domain. 
We validate our approach qualitatively, by morphing the same face according to different ad categories, and quantitatively, using human judgments and classifier accuracy. 

Our method works as follows. We first train facial expression and facial attribute classifiers using existing datasets. We detect faces in ads and predict their attributes and expressions. Next, we train a conditional variational autoencoder (CVAE) on our dataset of ad faces. The model learns to reconstruct an ad face from a vector comprised of a learned latent representation, facial attributes, and facial expressions. At test time, we embed all ad faces into vector space using our encoder and then compute how faces differ in that space across ad topics. Using these per-topic learned differences, we transform embeddings of other ad faces into each ad topic. Finally, we use our decoder on the transformed embeddings to generate distinct faces across ad topics. We show examples of our transformations in Fig.\ \ref{fig:concept}.


Note that prior work has modeled the conceptual rhetoric that ads use to convey a message \cite{hussain2017automatic,ye2018advise}, but no work models the visual variance in the portrayal of the same object across different ad categories, nor attempts to generate such objects. 

The contributions of our work are three-fold:
\begin{itemize}[noitemsep,nolistsep]
    \item We propose the problem of studying what makes an object visually persuasive and generating objects which convey appropriate visual rhetoric for a given ad topic.
    \item We analyze object frequency and appearance in ads, and discover objects with class-dependent appearances, which we then generate with promising quality.
    \item We develop a novel generative approach for modifying the appearance of faces into different ad categories, by elevating visual variance to a semantic level without the need for new semantic labels. The benefit of doing so is we can leverage semantics learned on larger datasets. Rather than directly modeling how faces in different ad categories differ on the pixel level, we model how they differ in terms of predicted attributes and facial expressions, then use these distinctions to create faces appropriate for a given ad category. Our method outperforms relevant baselines at this task.
\end{itemize}
\vspace{-0.5em}
\section{Related Work}
\label{sec:related}
\vspace{-0.5em}
\emph{Generative models.} Recently, generative adversarial networks (GANs) have produced impressive results on a number of image generation tasks \cite{choi2017stargan, berthelot2017began, zhu2017unpaired, ding2017exprgan, gurumurthy2017deligan}.  Conditional GAN variants have also been proposed which generate images subject to conditional constraints. \ct{For example, recent work by \cite{choi2017stargan, lample2017fader}} allows conditional transformations on specific images. Unfortunately, GANs are notoriously challenging to train, particularly on small and diverse datasets \cite{gurumurthy2017deligan}. Our autoencoder-based method is able to contend with such a dataset, while modeling meaningful differences across ad topics. \ct{Autoencoders \cite{hinton2006reducing, walker2016uncertain, pu2016variational, makhzani2015adversarial, lample2017fader, yan2016attribute2image}} are an older type of generative model, but perform competitively when trained with recent perceptual loss functions \cite{hou2017deep}. Autoencoders learn to project an image into a learned embedding space and then to reconstruct the original image from the embedding. Our method is a conditional variational autoencoder (CVAE) \cite{sohn2015learning}, which adds supervised information about ad faces into the model, along with custom loss functions we found to improve result quality.

\emph{Facial expressions and attributes.}
We condition our model on facial expressions recognized in faces for different ad categories. Facial expression and emotion recognition is an established and popular topic \cite{essa1997coding,kanade2000comprehensive,shan2009facial,liu2014facial,mollahosseini2017affectnet}. 
Usually seven canonical expressions are recognized: happiness, sadness, surprise, fear, disgust, anger, and contempt.
We also condition our generative model on facial attributes we predict on faces from ads. Attributes are semantic visual properties like ``bald,'' ``rosy cheeks,'' ``smiling'' or ``attractive'' \cite{liu2015faceattributes,lampert2014attribute,farhadi2009describing}. 

\emph{Visual persuasion.}
The primary novelty of our work is to discover what makes objects in ads persuasive and then to generate such objects. 
While no work has been performed in this space before, researchers have studied related problems. \cite{joo2014visual} learned to predict whether a photograph portrays a politician in a positive or negative light,
and \cite{joo2015automated} trained classifiers to predict the outcomes of elections based on the candidates' faces, but neither of these works creates a generative model. \cite{hussain2017automatic} propose a dataset of advertisements, and predict what message the ad conveys (e.g. ``buy this car because it is spacious'') but they do not model or generate the visual appearance of the same object across ad topics. 


\vspace{-0.5em}
\section{Approach}
\label{sec:approach}
\vspace{-0.5em}

We begin by describing how we extract faces from ads. We then describe how we predict attributes and facial expressions on the detected faces. Next, we present our autoencoder architecture and then describe how we use it to transform faces across ad categories.

\vspace{-1em}
\subsection{Ads data} 
\label{sec:data}
\vspace{-.25em}

We focus on the Ads Dataset of \cite{hussain2017automatic}. It contains ads belonging to 38 topic categories: beauty, soda, restaurants, etc. (called product ads) and domestic violence, safety, etc. (called public service announcements, or PSAs). We chose to study the ten most frequent product topics in the dataset, as well as all PSA topics, resulting in a set of 17 ad topics. 

\vspace{-1em}
\subsection{Face detection on ads}
\label{sec:approach:face_extraction}
\vspace{-.25em}

\begin{figure}[t]
  \centering
\includegraphics[width=1\textwidth]{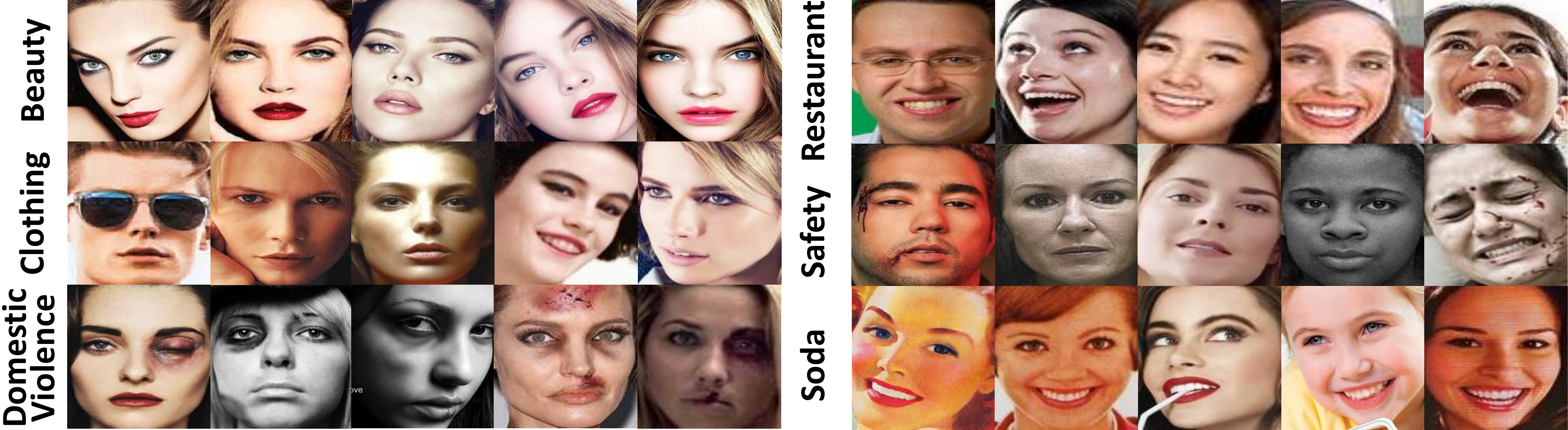}
\vspace{-2em}
\caption{We show examples of real faces from different categories of ads. We notice significant differences, many of which can be captured through facial attributes and expressions. }
\label{fig:faces}
\vspace{-1em}
\end{figure}

Our first step 
is to extract faces from ads. The remaining steps of our model work on this dataset of ad faces, rather than operating on whole ads images. This allows our model to concentrate on modeling and modifying facial appearance, without having to reconstruct the entire ad. 
We train Faster-RCNN \cite{ren2015faster} on the Wider Face dataset \cite{yang2016wider}. We remove face detections whose confidence is less than 0.85 or whose width or height is less than 60 pixels. 
We show examples of detected faces in different ad categories in Fig.~\ref{fig:faces}. In total, we detected 20,532 faces.  We observe, for example, that beauty ads often have brighter skin tones and feature women wearing makeup. Domestic violence faces are often darker and not smiling. Many soda faces appear vintage and smiling. Clothing ads are similar to beauty, but don't feature as bright of skin or makeup. Finally, safety ads feature more men and are not as dark as domestic violence ads. Importantly, many of the differences we observe are captured by facial attributes and expressions datasets.

\vspace{-1em}
\subsection{Predicting facial attributes and expressions}
\label{sec:approach:attributes_expressions}
\vspace{-0.25em}

We want our method to model the most relevant characteristics of faces in each ad topic category. 
As we observed in Fig.~\ref{fig:faces}, the differences between faces in different ad categories can naturally be described in terms of facial attributes and expressions. 
Because our dataset is small and diverse, our model may not have enough signal to reliably learn to model facial attributes and expressions without explicitly being directed to do so. 
In other words, it may devote its modeling power to matching the precise vintage or cartoon appearance of ad faces (i.e.\ low-level details) without learning a high-level model of recognizable semantic differences. 
Thus, rather than formulating our task as modeling the unconstrained distribution of pixels from the faces in each ad group, we manually inject high-level knowledge to facilitate manipulation of specific semantic attributes and expressions across ad topics.

We use the CelebA dataset \cite{liu2015faceattributes} of 40 facial attributes and the AffectNet dataset \cite{mollahosseini2017affectnet} of eight facial expressions plus valence and arousal scores. We train Inception-v3 \cite{szegedy2016rethinking} on each dataset. We train each classifier using a cross-entropy loss for classification. For the network trained on expressions, we add an additional classifier for the regression task of predicting the valence and arousal of the facial expression and also use a mean-squared error loss. 

Formally, let $\mathbf{I}_t$ represent the dataset of ad faces extracted from each ad topic $t$ (e.g. beauty faces, domestic violence faces, etc.). We use our trained attributes and expressions classifiers to predict these properties on our entire ad faces dataset. This results in an automatically labeled ads face dataset $\mathbf{I}_t = \left\{ \mathbf{x}^i_t, \mathbf{y}^i_t \right\}_{i=1}^{N_t}$, where $\mathbf{x}^i_t$ represents face $i$ from ad topic $t$, $\mathbf{y}^i_t$ represents the image's associated 50-dimensional vector (composed of 40 facial attributes and eight facial expressions with their accompanying valence and arousal scores), and $N_t$ represents the total number of faces per topic. We binarize our facial attribute predictions and represent our facial expressions in a one-hot fashion. The valence and arousal scores are real numbers from $[-1,1]$.
See our \fnurl{supplemental file}{http://www.cs.pitt.edu/~chris/files/2018/BMVC_2018_SUPP.zip} for these predictions for each ad topic.

\vspace{-1em}
\subsection{Conditional variational autoencoder}
\label{sec:approach:cvae}
\vspace{-0.25em}

Given an image $\mathbf{x}^i_t$ and conditional vector $\widehat{\mathbf{y}^i_t}$, which may differ from the image's ground truth signature, we seek a model $\mathbf{\theta}$ parameterizing the following transformation function:
\begin{equation}
f_{\mathbf{\theta}}\left(\mathbf{x}^i_t, \widehat{\mathbf{y}^i_t}\right)= \widehat{\mathbf{x}^i_t}
\end{equation}
where $\widehat{\mathbf{x}^i_t}$ is a face retaining the overall appearance of $\mathbf{x}^i_t$, but now bearing the attributes and expressions encoded in $\widehat{\mathbf{y}^i_t}$. If $\mathbf{y}^i_t=\widehat{\mathbf{y}^i_t}$, we  seek an unmodified reconstruction of $\mathbf{x}^i_t$. To modify the original appearance, we would like the reconstructed face to bear the provided set of attributes. If we denote our attribute and expression classifiers from Sec.\ \ref{sec:approach:attributes_expressions} jointly as $C$, we wish to enforce the following constraint:
\vspace{-0.1cm}
\begin{equation}
    C\left(f_{\mathbf{\theta}}\left(\mathbf{x}^i_t, \widehat{\mathbf{y}^i_t}\right)\right) = \widehat{\mathbf{y}^i_t}
\vspace{-0.1cm}
\end{equation} 
Thus, any modifications done by our model should result in our classifiers producing the same conditional vector that was provided to the transformation model.

We also seek the capability of transforming ad topic-wise facial appearance \textit{beyond} what is captured by our conditional vector. 
For example, if one topic features a predominant ethnicity, we would like our model to be capable of transforming a face into that ethnicity, even though it is not presented in our conditional vector. 
We thus seek a model capable of learning latent facial appearance information from our dataset. Autoencoders, which project an image into a low-dimensional space and then learn to reconstruct it from the sparse representation, are a natural choice. However, because we wish to interpolate faces across ad topics, enforcing that the learned space is smooth 
is important. We thus propose a custom conditional variational autoencoder, which enforces a Gaussian prior on the latent space \cite{sohn2015learning}. 

\begin{figure}[t]
  \centering
\includegraphics[width=1\textwidth]{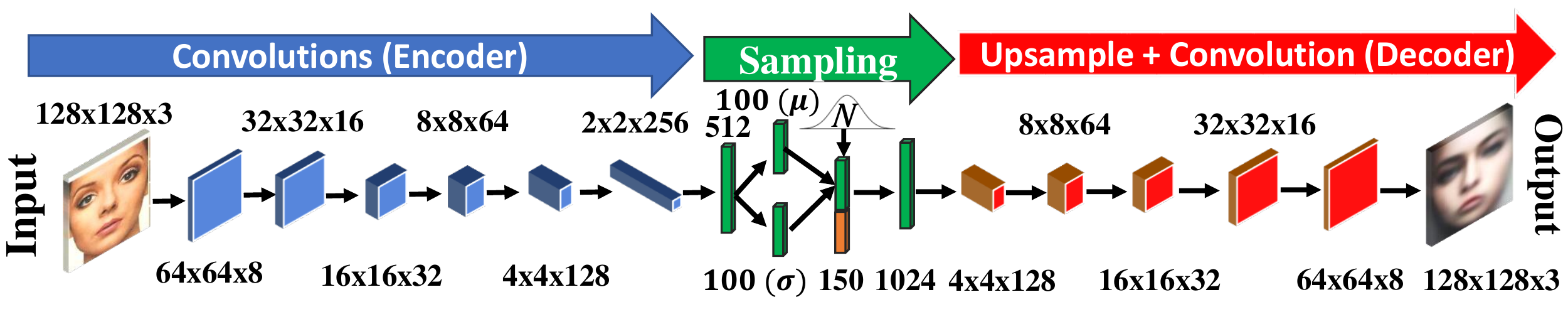}
\vspace{-2em}
\caption{We show our model transforming a beauty ad face into a domestic violence face. The conditional vector (orange bar) is appended to the sampled latent vector (see Eqs.~\ref{eq:latent},\ref{eq:conditional_and_latent}).}
\label{fig:architecture}
\vspace{-0.5em}
\end{figure}

We present our model's architecture in Fig.\ \ref{fig:architecture}. It contains two distinct components, an encoder and decoder, which are trained end-to-end to reconstruct ad faces. 
\vspace{-0.5cm}
\paragraph{Encoder.} Our encoder $g_\phi$ encodes any image $\mathbf{x}$ into the latent space $\mathbf{z}$ as follows:
\begin{equation}
\label{eq:latent}
\mathbf{z} = g_\phi(\mathbf{x}, \epsilon), \epsilon \sim \mathcal{N}
\end{equation}
where $\epsilon$ represents a vector sampled at random from $\mathcal{N}$, a standard normal distribution. Specifically, $g_\phi$ encodes an image by predicting $\mu$ and $\sigma$ for each dimension of the latent space. The latent embedding for an image is produced by combining $\epsilon$ with the predicted latent distribution parameters as follows:
$\mathbf{z} = \mu + e^{\frac{\sigma}{2}} \epsilon $.
This mechanism of predicting the latent variable (coupled with the smoothness constraint discussed later) represents an image as a sample drawn from a Gaussian image space. Thus, the same image's latent embedding will differ each forward pass of the encoder due to random sampling of $\epsilon$. This exposes our decoder network to a degree of local variation because
the decoder learns that a larger space of embeddings map to the same face. This encourages smoothness in the latent space, which is important for the interpolation on latent vectors performed later.

\vspace{-0.5cm}
\paragraph{Decoder.} 
We concatenate each image's latent vector with its associated conditional vector (attributes and expressions) to produce the final representation given to our decoder $p_\psi$: 
\begin{equation}
\label{eq:conditional_and_latent}
\mathbf{q}^i_t = \left[\widehat{\mathbf{y}^i_t}, \mathbf{z}^i_t\right] = \left[\mathbf{y}^i_t, g_\phi\left(\mathbf{x}^i_t, \epsilon\right)\right]
\end{equation}
During training, $\widehat{\mathbf{y}^i_t}=\mathbf{y}^i_t$. Our decoder network learns to reconstruct the original image from the embedding:
\begin{equation}
\widehat{\mathbf{x}^i_t} = p_\psi\left(\mathbf{q}^i_t\right) = p_\psi\left(\left[\mathbf{y}^i_t, g_\phi\left(\mathbf{x}^i_t, \epsilon\right)\right]\right)
\end{equation}

\paragraph{Learning.}
We train our model end-to-end to reconstruct the image provided to the encoder. However, because L2 reconstruction losses have been shown to produce blurry predictions \cite{mathieu2015deep}, we instead use a perceptual loss similar to \cite{hou2017deep}. Rather than compute the distance between the reconstruction and original image in pixel space, we compute the distance in \textit{feature space} of a pretrained VGG classification network following \cite{zhang2018unreasonable}. In our experiments, using a perceptual loss substantially improved the quality of reconstructions. Formally, let $\Phi\left(\mathbf{x}^i_t\right)$ and $\Phi\left(\widehat{\mathbf{x}^i_t}\right)$ represent the activations of layer \texttt{relu2\_2} of a pretrained VGG-19 \cite{simonyan2014very} network on the original and reconstructed images. The reconstruction loss $\mathcal{L}_r$ is given by:
\begin{equation}
\label{eq:reconstruction}
\mathcal{L}_r =  \left\lVert \Phi\left(\mathbf{x}^i_t\right) - \Phi\left(\widehat{\mathbf{x}^i_t}\right) \right\rVert_2^2 
\end{equation}

We provide our decoder with the predicted facial attributes and expressions $\mathbf{y}^i_t$ so that we know these aspects of faces will be represented and thus modifiable across ad categories. However, 
the decoder might ignore less conspicuous attributes, so we force it to use the conditional information. 
The model should produce samples that cause our classification networks to output the same vectors provided to the decoder. If $C_a$ and $C_e$ represent attribute and facial expression classifiers, our conditional classification loss $\mathcal{L}_c$ is given by:
\begin{equation}
\label{eq:conditional_classification}
\mathcal{L}_c =  l_{bce}\left(C_a\left(\mathbf{x}^i_t\right), C_a\left(\widehat{\mathbf{x}^i_t}\right)\right) + l_{nll}\left(C_{e_{exp}}\left(\mathbf{x}^i_t\right), C_{e_{exp}}\left(\widehat{\mathbf{x}^i_t}\right) \right) + l_2\left(C_{e_{va}}\left( \mathbf{x}^i_t \right),C_{e_{va}}\left( \widehat{\mathbf{x}^i_t} \right)\right)
\end{equation}
where $C_{e_{exp}}$ and $C_{e_{va}}$ represent the facial expression and valence and arousal predictions from $C_{e}$ respectively, $l_{bce}$ represents the binary cross entropy loss, $l_{nll}$ represents the negative log-likelihood loss after softmax is applied to the inputs (for multiclass classification), and $l_2$ represents the $l_2$ loss (for regression). In practice, we found our classification constraint improved reconstructions and made them more responsive to changes in the conditional vector. 

To encourage smoothness in the latent space, we use a standard KL divergence term which measures the relative entropy between a spherical Gaussian distribution and the latent distribution \cite{sohn2015learning}. The KL term $\mathcal{L}_{KL}$ can be analytically integrated \cite{kingma2013auto} into a closed form equation as follows:
\begin{equation}
\mathcal{L}_{KL} = \frac{1}{2} \sum e^\sigma + \mu^2 -1 - \sigma
\end{equation}
We found the KL constraint critical to producing smooth faces.
Our final loss is:
\begin{equation}
\mathcal{L} = \alpha \mathcal{L}_r + \beta \mathcal{L}_c + \gamma \mathcal{L}_{KL}
\end{equation}
\ct{where $\alpha$, $\beta$, and $\gamma$ are hyperparameters weighting the contribution of each loss component.}

\vspace{-1em}
\subsection{Cross-category facial transformation}
\label{sec:approach:transformation}
\vspace{-0.25em}
We described how to reconstruct a face, using an encoder, decoder, and fixed attributes and expressions. We now define what we input to our decoder, to translate a face to an ad class.

Notice that our model never accesses the ad topic category each face comes from.
This is because the faces within topic categories are too varied for the model to make use of topic information. 
However, in order to transform faces so they appear to come from different topics, we first must learn how faces differ in each topic. 
We compute a vector for each ad topic, which, when added to an image's embedding, makes the reconstruction appear more appropriate for that topic. Specifically, we compute the \textit{topic transformation vector} $\mathbf{v}_t$ for each topic $t$ as follows, where the horizontal bar indicates computing the mean per dimension:
\begin{equation}
\mathbf{v}_t = \sum^{N_t}_i \overline{\mathbf{q}^i_t} - \sum_{t^{\prime} \neq t} \sum^{N_{t^\prime}}_i \overline{\mathbf{q}^i_{t^\prime}}
\end{equation}
In order to make the transformations more visible, we increase the magnitude of the vector by multiplying the conditional portion of $\mathbf{v}_t$ by 10 and the latent portion by 2.5. We found this visibly improved the distinctiveness across topic categories. To translate a face $\mathbf{x}$ into ad category $t^\prime$, we modify the embedding of $\mathbf{x}$ using $\mathbf{v}_{t^\prime}$ and then reconstruct it as follows:
\begin{equation}
\widehat{\mathbf{x}^i_{t \rightarrow t^\prime}} = p_\mathbf{\psi} \left(\mathbf{q}^i_t + \mathbf{v}_{t^\prime} \right)
\end{equation}
\vspace{-2em}
\subsection{Implementation details}
\label{sec:approach:implementation}
\vspace{-0.25em}
We train our encoder and decoder end-to-end, but we do not train the VGG-19 network. 
We train the two classification Inception networks offline, before training our autoencoder.
We train using the Adam optimizer \cite{kingma2014adam} with learning rate 5.0e-4. We use minibatch size of 32 and train for 200 epochs. To ensure robustness to the highly varied ads faces dataset, we perform aggressive data augmentation.
We randomly horizontally flip the training data and also randomly zoom into or out of the images. We then crop the zoomed images to 128x128. 
This allows our models to be less sensitive to facial alignment. \ct{We empirically found using 100 dimensions for $\mathbf{z}$ to work well. We set $\alpha=1$ and $\beta$ and $\gamma$ to $0.0001$; larger values caused poor reconstructions.} We use Xavier initialization \cite{glorot2010understanding} and leaky ReLU activation \cite{he2015delving} for inner layers with negative slope 0.01. We find using batch normalization \cite{ioffe2015batch} with eps 1e-4 helps stabilize training. We implement all components of our model in PyTorch \cite{paszke2017automatic}.






\vspace{-0.5em}
\section{Experimental Validation}
\label{sec:results}
\vspace{-0.25em}
We conduct our experiments on the image advertisement dataset of \cite{hussain2017automatic}. 
We initially sought to study general object appearance across ad topics, but our analysis below revealed faces were by far the most distinct object per topic. We thus focus primarily on modeling faces. 

\vspace{-0.5em}
\subsection{Objects in Ads}
\vspace{-0.25em}

We ran a 50 layer residual RetinaNet \cite{lin2017focal} trained on the COCO dataset \cite{lin2014microsoft} on all ads in the 17 ad topics defined in Sec.~\ref{sec:data}. We first studied the \textit{distributions} of objects across ad topics.
We found many object-topic correlations, e.g.\ cars are most frequent in car ads, bottles occur frequently in alcohol and soda ads, animals are often found in animal rights and environment ads, etc. Overall, we found that people tended to occur 13 times more frequently than the second most common object (car).
We next studied how objects' \textit{appearance} differed across ad topic categories. We extracted SIFT \cite{lowe2004distinctive} features for each object and computed BoW histograms with $k=100$. 
We then analyzed the ``visual distinctiveness'' of objects, by measuring how each object's appearance changed within and across ad topics. 
We found that cars are highly visually distinct in car ads. This makes sense because cars in car ads are the \textit{focus} of the ad, not just a background object. We also found dogs were distinct in animal rights ads, cell phones in electronics ads, cake and bowl in chocolate ads, and bottle in soda ads. We include complete object distribution and visual distinctiveness tables in our supplementary file. Faces were the single object category which occurred frequently enough across topics to train a model on, thus we primarily focus on modeling faces in this paper.

\vspace{-0.5em}
\subsection{Qualitative Results}
\vspace{-0.25em}
\begin{figure}[!t]
\vspace{-0.25em}
  \centering
\includegraphics[width=1\textwidth]{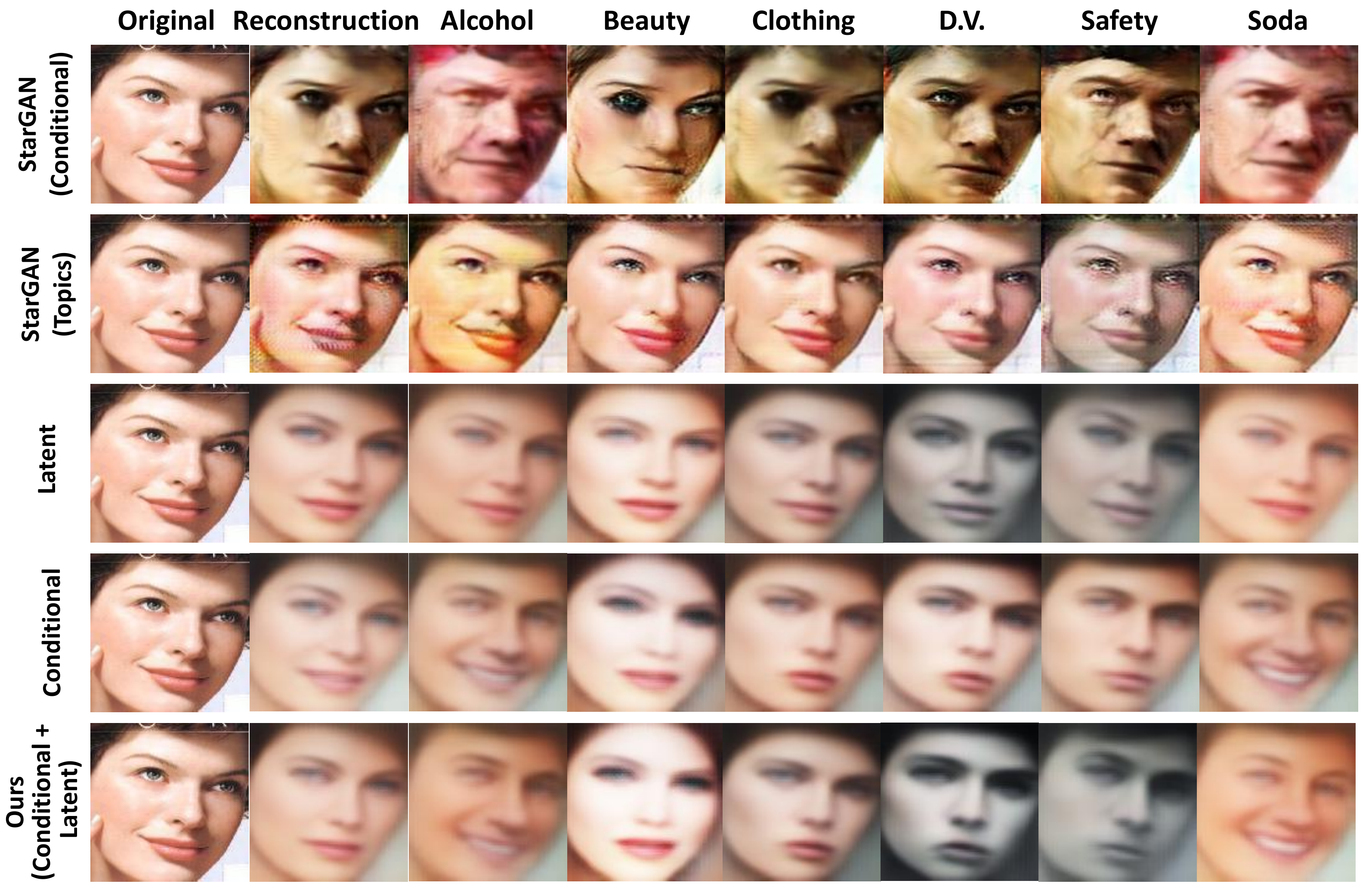}
\vspace{-2.4em}
\caption{We show the result of transforming the same face using five different methods. Our method (bottom row) most faithfully transfers the topic-specific facial appearance as judged by our human study. We include additional qualitative results in our supplementary.}
\label{fig:qualitative}
\vspace{-0.75em}
\end{figure}

\ct{We compare our method against two baselines inspired by attribute autoencoders \cite{hou2017deep, lample2017fader, yan2016attribute2image}, one of which has access to attributes and one which does not, as well as two variations of a state-of-the-art adversarial network for transforming images and attributes \cite{choi2017stargan}:}
\begin{itemize}[noitemsep,nolistsep]
\item \textbf{Conditional+Latent (Ours)} - Our full model, described in Sec.~\ref{sec:approach}.
\item \textbf{Conditional} - Our model trained with latent and conditional information (attributes, expressions, and valence/arousal), however only the 50 conditional dimensions are changed when translating a face across topics, while the latent dimensions stay fixed. 
\item \textbf{Latent} - Our model \emph{without} the conditioning on attributes and expressions.
\item \textbf{StarGAN \cite{choi2017stargan} (Conditional)} - We train StarGAN to modify faces to a given 50-dimensional conditional vector (facial attributes, expressions, valence/arousal).
\item \textbf{StarGAN \cite{choi2017stargan} (Topics)} - We train StarGAN to modify faces into a given topic. At training time, we train the model on the ground truth ad topic categories the faces are from. The model thus explicitly learns how facial appearance changes across topics.
\end{itemize}

In Fig.\ \ref{fig:qualitative}, we observe that our method \textbf{Conditional+Latent} produces the most noticeable and dramatic changes in visual appearance. We observe changes in gender, skin tone, facial expression, and facial shape. Alcohol ads feature smiling men, beauty ads tend to have light skin with lipstick, and clothing ads are similar, but with less skin brightness and less smiling. Faces in domestic violence ads are often frowning and darker, while those in safety ads tend to appear more masculine. Finally, soda ads have a vintage appearance with a large smile. For \textbf{Conditional}, we find that many aspects of the face change appropriately. However, the model is unable to transform other features not captured by the conditional vector: for example, making the face appear darker for domestic violence ads. 
For \textbf{Latent}, we find that while facial appearance overall changes, facial expressions and many facial attributes remain fixed, leaving a smiling face in inappropriate categories such as domestic violence. 

We observe that both versions of StarGAN maintain the original image's appearance, but do not change the image much per topic. We notice that \textbf{StarGAN (Conditional)} tends to produce smoother skin and highlighted eyes for ``beauty,'' but its other categories are harder to discern. \textbf{StarGAN (Topics)} adds low-level details into the generated images in order to achieve a lower topic prediction loss rather than changing the facial appearance of the image. 

\vspace{-0.5em}
\subsection{Quantitative Evaluation of Generated Faces}
\vspace{-0.25em}

In addition to our qualitative results, we perform two quantitative experiments to assess how well our method transforms faces into each ad topic. For our first experiment, we perform a human study to assess how well humans perceive each method to do in terms of data transformation. Eight non-author participants participated in our study.
We first show them examples of real faces from five ad categories: beauty, clothing, domestic violence, safety, and soda. To ensure our judges pay attention to the visual distinctions, we ask them to classify 10 rows of real faces into the correct ad topic. We then show participants the same image translated by five randomly sorted methods into the five topics, and ask them to select the method which best portrays the distinct visual appearance of faces across the five topics. 

For our second experiment, we transform the same faces into all 17 ad categories. Next, we finetune a pretrained AlexNet \cite{krizhevsky2012imagenet} on the transformed faces to predict which topic the face is supposed to portray. Finally, we evaluate our model on \textit{real} faces. Thus, methods which reliably transform faces in ways which capture the distinct traits of each topic will achieve higher classification accuracy. This metric assesses how well discriminative features are translated into generated ads but does not assess the visual quality of the results or the task we are ultimately interested in, namely producing visually distinct faces across topics.

\begin{table}[t]
\vspace{-0.25em}
\centering
\resizebox{1\linewidth}{!}{%
\begin{tabular}{|c|c|c|}
\hline
\textbf{Method} & \textbf{\begin{tabular}[c]{@{}c@{}}Human Judgments:\\ Best At Topic Transformation\end{tabular}} & \textbf{\begin{tabular}[c]{@{}c@{}}Classifier Topic \\ Prediction Accuracy\end{tabular}} \\ \hline
\textbf{StarGAN (Conditional)} & 0.100 &  0.069 \\ \hline
\textbf{StarGAN (Topics)} & 0.113  & \textbf{0.100} \\ \hline
\textbf{Latent} & 0.038 & 0.086 \\ \hline
\textbf{Conditional} & 0.144 & 0.080 \\ \hline
\textbf{Conditional + Latent (Ours)} & \textbf{0.606}  & 0.092 \\ \hline
\end{tabular}%
}
\vspace{-0.5em}
\caption{We present two quantitative results. The first shows in what fraction of examples humans chose each method as the best, for generating visually distinct and appropriate faces in each topic. The rightmost column shows the accuracy of a classifier when trained on each row's synthetic training data and tested on real images from 17 categories.}
\label{table:quantitative}
\vspace{-0.75em}
\end{table}


We present quantitative results in Table \ref{table:quantitative}. \textbf{Ours} performs best at the objective we set out to accomplish, and does competitively on the objective but less informative classifier accuracy task. In our human study, human judges found that our method best generates topic-specific faces nearly \textbf{\textit{4 times as often}} as the next best method, \textbf{Conditional}. Interestingly, humans rarely prefer the \textbf{Latent} model, demonstrating the importance of including attributes and facial expressions. For the classification task, the classifier trained on \textbf{StarGAN (Topics)}'s data performs best, followed by our method. This makes sense because \textbf{StarGAN (Topics)} sees images labeled with topics at train time and learns what features are useful for topic classification. However, we see that the method only changes low-level details (e.g.\ color) without changing any semantics. Our method changes face semantics, never sees topic information at train time, yet performs on par with \textbf{StarGAN (Topics)} on this task, confirming our method does transfer topic-specific appearance. We observe that the accuracy for all models is similar and fairly low. This is most likely because many faces are impossible to classify since they are generic, non-persuasive background faces. Please see our supplementary file for additional results, including evaluation of visual quality and human classification performance.


\vspace{-0.5em} 
\subsection{Generating Other Objects}
\vspace{-0.25em}
\begin{figure}
\label{fig:began}
  \centering
\includegraphics[width=1\textwidth]{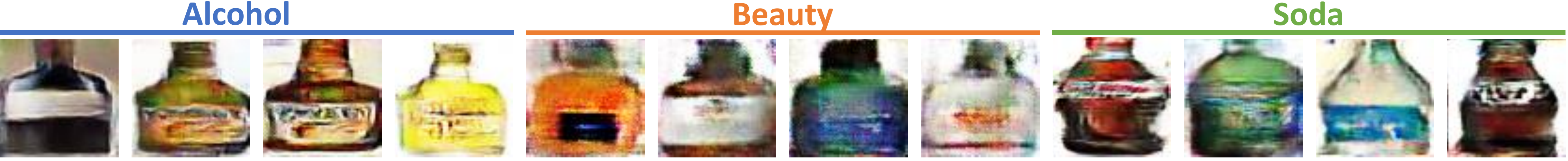}
\vspace{-2em}
\caption{Alcohol, beauty, and soda bottles generated using our implementation of a conditional BEGAN \cite{berthelot2017began} trained on bottles. We observe interesting differences across ad topics. }
\end{figure}
We wanted to see whether we could generate other objects besides faces as they appear in different ad categories. We conditioned BEGAN \cite{berthelot2017began} on ad topics and trained on bottles from alcohol, beauty, and soda ads. We used an image size of 64x64 due to the limited amount of training data per class. We observe that the model does learn meaningful topic-wise differences in object appearance. For example, alcohol bottles look like liquor bottles with a long stem, beauty bottles are wider with a short stem (perfume), soda bottles have a soda bottle shape and label. These results show that intra-topic object appearance can be modeled, but future work is needed to address problems such as mode-collapse.



\vspace{-1em}
\section{Conclusion}
\label{sec:conclusion}
\vspace{-0.25em}
In this paper, we studied how objects appear in different categories of ads and how ads use these objects for persuasion. Based on our object analysis, we focused on faces and explored how faces could be generated across different types of ads. We proposed a conditional variational autoencoder for this task, which we augment by providing high-level facial attributes and expressions; experiments showed this auxiliary supervision was critical to achieving good results. Our experiments confirm that our method greatly outperforms a variety of baselines. We also show early results on how topic-specific objects beyond faces may be generated. In future work, we will investigate techniques for reliably generating other objects,
and for creating ads complete with multiple objects and persuasive slogans.

\vspace{1em}
\noindent \ct{
\textbf{Acknowledgements:}
This material is based upon work supported by the National Science Foundation under Grant Number 1566270 and a NVIDIA hardware grant. Any opinions, findings, conclusions or recommendations expressed in this material are those of the authors and do not necessarily reflect the views of the National Science Foundation. We are grateful to our human evaluation study participants and the reviewers for their time and effort. Finally, we thank Sanchayan Sarkar for preparing the AffectNet data.
}
\vspace{-1.25em}

\bibliography{egbib}
\end{document}